\documentstyle[]{article}
\twocolumn

\begin{document}

\title{\large \sc Dilemma - An Instant Lexicographer}
\author{\normalsize
       {\sc Hans Karlgren, Jussi Karlgren,
            Magnus Nordstr\"om, Paul Pettersson,
            Bengt Wahrol\'en} \\
        \verb!dilemma@sics.se! \\[5pt]
	Swedish Institute of Computer Science \\
	Box 1263, 164 28 Kista, Stockholm, Sweden \\[5pt]
}

\maketitle
\thispagestyle{empty}

\vspace*{-5mm}
\begin{center} \subsection*{Introduction} \end{center} \noindent
Dilemma is intended to enhance quality and increase productivity of
expert human translators by presenting to the writer relevant lexical
information mechanically extracted from comparable existing
translations, thus replacing - or compensating for the absence of - a
lexicographer and stand-by terminologist rather than the translator.
Using statistics and crude surface analysis and a minimum of prior
information, Dilemma identifies instances and suggests their
counterparts in parallel source and target texts, on all levels down
to individual words.

Dilemma forms part of a tool kit for translation where focus is on
text structure and over-all consistency in large text volumes rather
than on framing sentences, on interaction between many actors in a
large project rather than on retrieval of machine-stored data and on
decision making rather than on application of given rules.

In particular, the system has been tuned to the needs of the ongoing
translation of European Community legislation into the languages of
candidate member countries. The system has been demonstrated to and
used by professional translators with promising results.

\vspace*{-2mm}
\begin{center} \subsection*{Instant Lexicographer} \end{center} \noindent
The design of translation aids beyond ordinary text processing and
database accession and maintenance tools is mostly based on the same
simplified view which --- for compelling reasons --- has been the
working hypothesis of machine translation: that the source text has a
well-determined meaning and that there exists in the target language
at least one correct and adequate ways of expressing that meaning.

When these assumptions are reasonably well
justified, translation is relatively easy, fast and cheap with
traditional methods and mechanization not rarely feasible with
methods now known or envisaged. Typically, however, the translator
must do more than retrieve and operate on pre-established and in
principle pre-storable correspondences. Thus, lexical correspondences
do not exist for all items; it is an essential part of translation to
establish them. Legal texts, factual and stereotype though they may
seem, regularly represent thoughts, attitudes and arguments which do
not have any counterparts in the target language prior to
translation. This is particulary true in the huge project to
translate the European Community legislation into the languages of
countries which are not yet members of the Community and which
currently have a partly different legal conceptual framework.

What human translators need is decision support.
The most important tools are telephones, electronical conferencing
systems and good and relevant dictionaries.  Unfortunately, there are
not always at every point of time knowledgable and cooperative
colleagues or other experts to call, electronical networks are only
recently being established in some domains, and the intelligent and
comprehensive dictionaries, which can serve as a writer's digest to
the cumulative literature in a field are few and far between. Answers
are often to be found in a text translated late at night the day
before - or in the preceding sections of the text at hand. Rather
than an automated writer, we need an instant lexicographer.

\vspace*{-2mm}
\begin{center} \subsection*{Recycling Translations} \end{center} \noindent
In practice, existing translations are being used as a major source
(S{\aa}gvall Hein {\it et al}, 1990; Merkel 1993). Often in the hope
to be able to avoid duplication of costs - or of getting paid twice for the
same effort - by finding identical or near-identical texts or passages,
but, more importantly, to ensure consistency or getting good suggestions,
to follow or argue against. Synonymy variation for the same concept is not
appreciated in technical and legal prose and avoided as anxiously as
homonymy. The ideal is 1:1 correspondences between expressions at least
within one pair of documents and to eliminate ``forks'', i.e., one
expression being translated into or being the translation of more than one
counterpart in the other language (Karlgren, 1988).

We shall call a coupled pair of source and target text a {\em bitext}
(Isabelle, 1992). What is said here about bitexts can be generalized to
n-tuples of parallel texts, claimed to differ ``only'' in language. Such
n-tuples exist: in the European Community, a major part of the legislation
is available in 9 ``authentic'' versions, which in (legal and political)
theory are equivalent, and according to plans the number of ``authentic''
will soon rise to 12 or more. Little efforts have previously been made to
systematically exploit this redundancy by means of potent multi-lingual
procedures for retrieving facts or expressions, even when surprisingly
simple methods show promise of surprisingly useful results (Dahlqvist, 1994).

\vspace*{-2mm}
\begin{center} \subsection*{Steps in the Translation Process} \end{center}
\noindent
Producing target language text is only a small proportion of the
translation process. Empirically, good economy is achieved if about
the same proportion of work is put into each of the stages
Preparation, Text production and Verification, a trichotomy
reminiscent of the classical person-time breakdown of software
development (Brooks, 1975). The Dilemma tool is useful for {\em some} tasks
in {\em each} of the three stages.

\vspace*{-2mm}
\begin{center} \subsection*{Functionality} \end{center} \noindent
A typical question for translators while actually writing is
how a word or phrase has been used or translated in previously
processed texts. Conversely, they may ask for the source languages
counterparts of given target language expression, to make sure that
homonymy is not introduced. Similarly, during the preparation and
verification phases, a translator or editor scans through the text
for words and phrases that need to be resolved or treated specially.

\vspace*{-2mm}
\begin{center} \subsection*{Text Production Phase} \end{center} \noindent
\vspace*{-5mm}
\begin{center} \subsubsection*{Navigating in Bitexts} \end{center} \noindent
The first service is to enable the translator to browse through
the bitext and look at text elements
pairwise, to check for conventions of usage that are unfamiliar or
unexpected.

Pointing at a shorter or longer string in either language the user
can find successively larger contexts and their counterparts
or {\em countertexts} in the other language version. This service is
available to the user from within a word processor, the answer
presented in a separate window.

\vspace*{-5mm}
\begin{center} \subsubsection*{Counterwords} \end{center} \noindent
The second service is to assess the word-level counterparts
or ``counterwords'' so far used for a given word. Here the system
performs, crudely but instantly, the job of a terminologist or
lexicographer. It uses a statistical matching process which offers the
translator a list of candidate counterparts.

\vspace*{-2mm}
\begin{center} \subsection*{Verification phase} \end{center}
\vspace*{-5mm}
\begin{center} \subsubsection*{Translation Verification} \end{center} \noindent
In this phase a revisor reads the text to detect inadequacies and
inconsistencies. Often, there is no true answer to a terminological
question: either one of a few options may be equally good per se but
unintended variation is disturbing and may be misleading.
Verification, therefore, is not a matter of local correctness or of
compliance with a given dictionary or other norm, and reading one
passage at a time will not reveal the deficiency of the translation
(Karlgren, 1988).

One way of resolving or detecting dubious cases is to compare how a
word or phrase has been used in a multitude of previous contexts and
how it was rendered in their respective countertexts.

\vspace*{-2mm}
\begin{center} \subsection*{Preparation phase} \end{center} \noindent
\vspace*{-10mm}
\begin{center} \subsubsection*{Text-and Domain-specific Phrase Lists}
\end{center} \noindent
In the preparation phase the translator or editor has to
establish text- and domain-specific word and phrase lists. In a batch
mode, Dilemma produces draft lists on the basis of previously
translated material in the same domain.

\vspace*{-2mm}
\begin{center} \subsection*{Structuring Bitexts} \end{center} \noindent
For bitexts to be exploitable as information sources, text
constituents in the two versions must be paired on some hierarchical
levels - phrase, clause, sentence, paragraph, etc. We must create a
structured bitext, with links from each constituent not only to its
predecessor and successor but also to its counterpart in the other
language version. This cross-language structure can be rather easily
captured when the translation is being typed, but we need to be able
to derive the pairs from two given complete texts. Dilemma does so
automatically.

We make three linguistic assumptions:
\begin{enumerate}
\item The two texts can be segmented into hierarchical constituents so
      that most constituents in one text have a counterpart in the other.
\item For all levels, except the lowest level, counterparts occur in the
      same mutual order.
\item The counterparts on the lowest level, ``counterwords'', appear
      approximately in the same mutual order.
\end{enumerate}
\noindent
We do not assume every constituent on any level to have a counterpart,
nor constituents to be separated by unique delimiters.
Thus, if paragraphs are separated by a blank line and sentences by a
full stop followed by a space, we do not exclude that, say, a
paragraph in one language is sometimes rendered as an enumeration,
separated by blank lines and that ``1.5'' is now and then typed as
``1. 5''. The procedure is robust in that it tolerates gaps and none too
frequent deviations from the prevalent pattern.

We apply two statistical procedures, one of alignment for higher
levels and one of assignment for the lowest, ``phrase'', level.

\vspace*{-2mm}
\begin{center} \subsection*{Alignment} \end{center} \noindent
The general problem of order-preserving alignment on one linguistic
level reduces to the string correction problem (Wagner and Fischer,
1974). The practical solution is not trivial, however, due to the
extremely large search space even for small texts. We use an
algorithm with search space constraining heuristics not entirely
unlike the one published by Church and Gale (1990) but using
linguistic information on more levels. Using a minimum of prior
information, texts are aligned down to phrase level. Recognizing
identity or similarity of a few punctuation marks, numerals and the
number of words between these suffices for a crude alignment.

\vspace*{-2mm}
\begin{center} \subsection*{Word Assignment} \end{center} \noindent
When the two texts have been aligned on higher levels,
correspondences are established between
counterwords, which do not necessarily appear in the same order
in the two language versions. For this purpose Dilemma uses an
association function which is a weighted sum of measures of agreement
of word position within the phrase, relative frequency of occurrence,
and, optionally, some other properties. The weighting of the
parameters is set after text genre specific experimentation. Pairs of
terms with a high association value are candidate counterwords
(Nordstr\"om and Petterson, 1993).

The procedure is self-evaluating since uncertainty is reflected by a
low maximum association value. Only items which have a score above a
cut-off threshold are presented to the user.  The procedure yields
some 90 per cent successful assignments among those presented on the
basis of as little material as a single 10 page document, but for
rare words the assignment becomes less certain. In a material of 10
000 pages of legal documents related to the European Economic Space
as much as 50 per cent of the word tokens were hapax legomena and 75 per cent
occurred less than 5 times, providing a meagre basis for statistical
analysis.
These results can be improved if other properties are taken into
account. When a word length was included as a parameter in the
association evaluation, the results became marginally more adequate.
Syntactical tagging, vide infra, is expected to affect assignment
more.

\vspace*{-2mm}
\begin{center} \subsection*{Tagging} \end{center} \noindent
In the first release of Dilemma, alignment and assignment was
performed on unmodified typographical strings but naturally the
procedures were intended to be applied after monolingual
preprocessing. Trivially, results become practically much more
adequate and the statistical analysis more effective if, say, making
and made and the infinitive make are subsumed under one item and the
infinitive and the noun make are kept separate.  Without any change
of method, the procedure can be applied to strings of words tagged
morphologically and syntactically. The tools chosen for this purpose
are the parsers for English, French and Swedish developed at Helsinki
University (Voutilainen {\em et al}, 1993).

\vspace*{-2mm}
\begin{center} \subsection*{Implementational Status} \end{center} \noindent
Dilemma has been implemented in C++ and runs under
Microsoft Windows on a regular-size personal computer.
Dilemma is currently being evaluated and tested by
translators currently involved in the translation of large
amounts of legal documents into Scandinavian languages in
the context of the proposed accession to the European Economic Community.
\begin{center} \subsection*{References} \end{center}
{\small
\begin{description}

\item[Frederick Phillips Brooks.] 1975. {\em The mythical
      man-month -- essays on software engineering},
      Reading, Massachusetts: Addison-Wesley.

\item[Bengt Dahlquist.] 1994. {\it TSSA 2.0} A PC Program for Text Segmentation
      and Sorting, Department of Linguistics, Uppsala University, Uppsala.

\item[Gale,] {\bf W.A. and K.W. Church}. 1991.
      ``Identifying Word Correspondences in Parallel Texts'',
      in {\em Proceedings of the 4th Speech and
	Natural Language Workshop}, DARPA, Morgan Kaufmann.

\item[Brian Harris.] 1992. ``Bitext'',
     in {\em Proceedings of ``Trans\-lation and the European
     Communities''}, Biskops-Arn\"o, Stockholm:
     FAT (The Swedish Association for Authorized
     Translators).

\item[Pierre Isabelle.] 1992. ``Bitexts: Aids for
      Translators'',  {\em Screening Words: User Interface for
      Text}, 8th Annual Conference of the UW centre of
      the New OED and Text Research, Waterloo,
      Canada: Waterloo University.

\item[Hans Karlgren.] 1987. ``Making Good Use of Poor
      Translations'', in {\sc International Forum On
      Information And Documentation}, 12:4, Moscow: FID.

\item[Hans Karlgren.] 1988. ``Term-Tuning, a Method for
       the Computer-Aided Revision of Multi-Lingual
       Texts'', in {\sc International Forum On
       Information And Documentation}, 13:2, Moscow: FID.

\item[Hans Karlgren.] 1981. ``Computer Aids in Translation'' with the
      Hanzinelle Declaration, in Sigurd and Svartvik (eds.),
      {\it AILA Proceedings}, pp 86-101, Lund: AILA.


\item[Martin Kay.] 1980. ``The Proper Place of Men and
      Machines in Language Translation'', {\it Xerox report
      CSL-80-11}, Palo Alto: Xerox Palo Alto Research Center.

\item[Magnus Merkel] 1993. ``When and Why Should Translations be Reused?'',
     {\it Papers from the 18th UAAKI Symposium on LSP, Theory of
     Translation and Computers}, V\"oyri.

\item[Magnus Nordstr\"om and Paul Petterson.] 1993. ``A
      Tool for Rapid Manual Translation'', {\it Master's
      Thesis at the University of Uppsala}, Uppsala:University of Uppsala.

\item[Anna S{\aa}gvall Hein], {\bf Annette \"Ostling, Eva Wikholm}. 1990.
``Phrases in the Core Vocabulary''. {\it A Report from The Project Multilingual
Support for Translation and Writing}. Report no. UCDL-R-90-1. Center for
Computational Linguistics. Uppsala University.

\item[Atro Voutilainen and Pasi Tapanainen.] 1993.
      ``Ambiguity Resolution in a Reductionistic Parser'',
      {\it Proceedings of the 6th Conference of the European Chapter of the
      ACL}, pp. 394-403, Utrecht:ACL.
\item[Robert A. Wagner and M. J. Fischer.] 1974. ``The
      String-to-String Correction Problem'',
      {\sc Journal Of The ACM}, 21:1, pp 168-173, New York:ACM.
\end{description}
}
\end{document}